\documentclass[runningheads]{llncs}
\usepackage[T1]{fontenc}
\usepackage{graphicx}
\usepackage{color}
\usepackage{amsmath,graphicx,color,hyperref}
\usepackage{multirow}
\usepackage{wrapfig}
\usepackage{amssymb}
\usepackage{cite}
\usepackage{makecell} 

\newcommand{\dd}{\scriptsize{\textnormal{d}}}

\DeclareRobustCommand{\ShowColormap}{\raisebox{-0.14em}{\includegraphics[height=.8em]{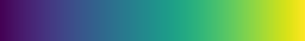}}}

\begin{document}
\title{Predicting Shape Development: A Riemannian Method}
%
%
\author{Do\u ga T\"urkseven\inst{1} \and
Islem Rekik\inst{1,2}\orcidID{0000-0001-5595-6673} \and
Christoph von Tycowicz\inst{4}\orcidID{0000-0002-1447-4069} \and
Martin Hanik\inst{3,4}\orcidID{0000-0002-7120-4081}
}
\authorrunning{D. T\"urkseven et al.}
%
%
\institute{BASIRA Lab, Istanbul Technical University, Istanbul, Turkey \and
Computing, I-X and Department of Computing, Imperial College London, London, UK \\
\email{i.rekik@imperial.ac.uk}\\ \and
Freie Universit\"at Berlin, Berlin, Germany\\ \and
Zuse Institute Berlin, Berlin, Germany \\
\email{\{hanik,vontycowicz\}@zib.de}}
%
\maketitle              
\begin{abstract}
    Predicting the future development of an anatomical shape from a single baseline observation is a challenging task. But it can be essential for clinical decision-making. Research has shown that it should be tackled in curved shape spaces, as (e.g., disease-related) shape changes frequently expose nonlinear characteristics. We thus propose a novel prediction method that encodes the whole shape in a Riemannian shape space. It then learns a simple prediction technique founded on hierarchical statistical modeling of longitudinal training data. When applied to predict the future development of the shape of the right hippocampus under Alzheimer's disease and to human body motion, it outperforms deep learning-supported variants as well as state-of-the-art. 

\end{abstract}
\section{Introduction} \label{sec:intro}


Shapes of anatomical structures are of considerable medical interest, and they are encountered particularly often in the analysis of medical images. Studies have shown that they should be modeled as elements of curved manifolds---shape spaces---instead of ordinary Euclidean space~\cite{PennecSommerFletcher2020}. Thus, it is imperative to develop methods for such spaces when working on problems involving anatomical shapes.  


A particularly interesting and relevant task is the prediction of the future development of a shape---``How will an anatomical structure look like after a certain amount of time has passed?''
This question is of great interest as shapes of anatomical structures are often correlated with (states of) diseases; see, e.g.,~\cite{PennecSommerFletcher2020,caliva2022boneshapeaging,Navayazdani_ea2020}.
Predicting how a shape will develop in the future could thus play a significant role in diagnosis and prevention as well as aid physicians in their choice of treatment. 

One important example is the relation between the shape of the hippocampus and Alzheimer's disease, as previous studies have shown that the former can be used to discriminate between Alzheimer's and normal aging~\cite{gerardin2009classification}.
Thus, if the longitudinal development of the hippocampus' shape can be predicted, a better prognosis can be achieved; and with Alzheimer's improved early diagnosis can have a serious positive impact~\cite{rasmussen2019earlydiagnosis}. 


To uncover developmental trends in populations, longitudinal studies that involve repeated observations of individuals play an essential role.
The variability in such data can be distinguished as cross-sectional (i.e., between individuals) and longitudinal (i.e., within a single individual over time).
The latter is highly correlated, violating the independence assumption of standard statistical tools like mean-variance analysis and regression, thus requiring inferential approaches that can disentangle cross-sectional and longitudinal effects.
For shape-valued data, another challenge that warrants attention is that curved spaces lack a global system of coordinates.
Assessing differences in longitudinal trends requires a notion of transport between tangent spaces to spatially align subject-specific trajectories.
For manifolds, \textit{parallel transport} has been shown to provide highly consistent transports~\cite{lorenzi2013geodesics, Navayazdani_ea2020} with improved sensitivity over other methods. Specifically, in~\cite{bone2017prediction} the adequateness of geodesic subject-wise models for the progression of subcortical brain
structures in Alzheimer's disease and the potential for prognosis via parallel transport of individual trends has been shown.


To assess longitudinal and cross-sectional shape variation jointly, hierarchical statistical models pose an adequate and very flexible framework~\cite{GFS2016}.
In recent years, various generalizations of such models to manifold-valued data have been proposed based on probabilistic~\cite{bone2020learning} and least-squares theoretic~\cite{MF2012,nava2022hierarchical,hanik2022nonlinear} formulations.
These approaches account for the inherent interrelations by describing each subject with its own parametric spatiotemporal model---most prominently geodesics.
Additionally, the subject-specific trends are assumed to be perturbations of a population-average trend, which is referred to as the ``fixed'' effect and is often of primary interest.


Population-level analysis apart, a few studies have focused on shape development prediction \textit{for individuals}.
A deep learning pipeline for predicting longitudinal bone shape changes in the femur to diagnose knee osteoarthritis was introduced in~\cite{caliva2022boneshapeaging}. This approach utilizes a spherical encoding to map a 3D point cloud of the bone into a 2D image. It thereby relies on the assumption that the femora are (approximately) star-shaped. While the proposed pipeline produces accurate predictions of future bone shape, anatomies in general (e.g., hippocampi) are not star-shaped prohibiting a spherical encoding. A further limitation of the method is that it needs 3 separate observations for its prediction---a requirement that strongly hinders early diagnosis.

A varifold-based learning approach for predicting infant cortical surface development has been proposed in~\cite{rekik2016predicting}. The method uses regression on varifold representations~\cite{fishbaugh2017geodesic} to learn typical shape changes and uses the latter to predict from a single baseline. Although the method partly uses the curved space of diffeomorphisms, it only employs it for pre- and post-processing. It further requires the user to set several data-dependent parameters as optimal as possible (i.e., two data-dependent kernel sizes, a weighting between terms in the loss of the regression, the concentration of time points, and the number of neighbors to be considered in a nearest neighbor search), leading to a relatively high entry-barrier for users.

In this work, we propose a novel method based on hierarchical statistical modeling to predict shape evolution.
In contrast to previous geodesic approaches~\cite{bone2017prediction} that relied on hand-picked reference individuals, we provide a data-driven approach for learning shape progression.
Representing subject-wise trends as geodesics in shape space allows us to learn from longitudinal observations while respecting within-subject correlations. 
After training on whole trajectories, the prediction requires only a single shape making it applicable when only the baseline observation of the longitudinal development is given. 
Conceptually, our prediction is very simple as it only consists of a parallel translation of the initial velocity of the mean trend and a subsequent evaluation of a geodesic. It thus provides a high degree of interpretability.
Furthermore, no data-dependent parameters need to be set, which makes the method very user-friendly.

To validate our approach, we apply it to two real-world problems. We use it to predict the development of the shape of the right hippocampus under Alzheimer's.
To the best of our knowledge, this is the first approach that applies longitudinal statistical modeling to the prediction of shape developments from a single baseline.
We also use our method to predict human body motion as an example of an application in which large shape changes occur.
In both applications, we outperform state-of-the-art by a considerable margin. Our method even slightly outperforms deep-learning-enriched variants.
Upon acceptance of the paper, we will make our source code public in order to make our method easily accessible.

\section{Method}

\begin{figure*}[t]
    \begin{center}
        \includegraphics[width=\textwidth]{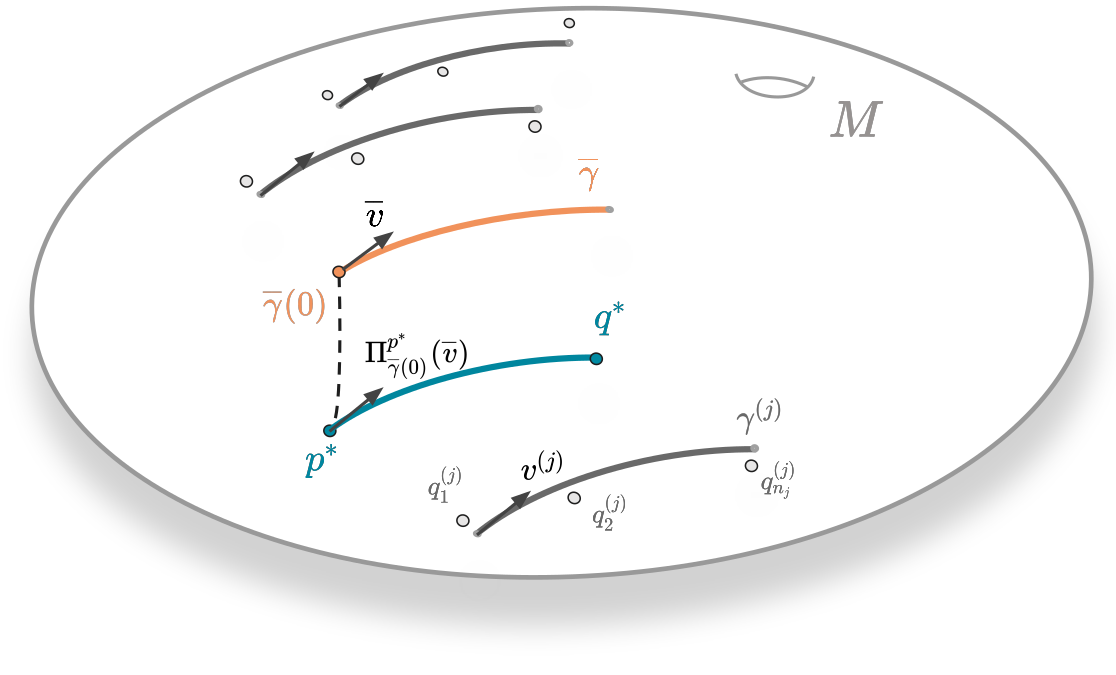}
        \caption{Depiction of the shape prediction method in a shape space $M$.}
        \label{fig:Model}
    \end{center}
\end{figure*}


    We first recall the necessary basics from Riemannian geometry and geometric statistics; a good reference on them is~\cite{PennecSommerFletcher2020}.

    A Riemannian manifold is a differentiable manifold $M$ together with a Riemannian metric $\langle \cdot,\cdot \rangle_p$ that assigns a smoothly\footnote{Whenever we say ``smooth'' we mean ``infinitely often differentiable''.} varying scalar product to every tangent space $T_pM$. The metric also yields a (geodesic) distance function $d$ on $M$.
    A further central object is the Levi-Civita connection $\nabla$ of $M$. Given two vector fields $X, Y$ on $M$, it is used to differentiate $Y$ along $X$; the result is again a vector field, which we denote by $\nabla_X Y$.
    With a connection one can define a geodesic $\gamma$ as a curve without acceleration, i.e., $\nabla_{\gamma'}\gamma' = 0$, where $\gamma' := \frac{\dd}{\dd t} \gamma$.
    It is a fundamental fact that each element of $M$ has a so-called normal convex neighborhood $U$ in which any two points $p,q \in U$ can be joined by a unique length-minimizing geodesic $[0,1] \ni t \mapsto \gamma(t;p,q)$ that does not leave $U$. Since it is the solution of a second-order differential equation, $p$, and $\gamma'(0)$ determine $\gamma$ completely. Indeed, for every $p \in U$ there are the Riemannian exponential $\exp_p: T_pM \to U$ and logarithm $\log_p := \exp^{-1}: U \to T_pM$ with $\exp(v) := q$ such that $\gamma'(0; p, q) = v$ and $\log_p(q) = \gamma'(0; p, q)$. 
    
    A further important fact is that in a Riemannian manifold, one usually cannot identify tangent spaces with each other. Therefore, tangent vectors must be transported explicitly along curves between points---the so-called parallel transport. This process depends on the chosen path; however, in $U$, we can always transport along the geodesic that connects the origin and destination. Therefore, whenever we speak of parallel-translating a vector $v$ from some $p \in U$ to $q \in U$ transport along the geodesic from $p$ to $q$ is meant; the resulting vector (which is in $T_qM$) is denoted by $\Pi_p^q(v)$. 
    
    Whenever we work with data, means are of interest. Given $q_1,\dots,q_n \in M$, their Fr\'{e}chet mean is the minimizer of the Fr\'{e}chet variance $F(p) := \sum_{i=1}^n d(p,q_i)^2$. 
    
    Interestingly, the set of all geodesics $G(U):= \{\gamma:[0,1] \to U\ |\ \gamma \textnormal{ geodesic} \}$ in $U$ can also be given the structure of a Riemannian manifold imposing a functional-based Riemannian metric~\cite{nava2022hierarchical}. As a consequence, we can compute Fr\'{e}chet means of geodesics in $G(U)$ if they are sufficiently localized; we assume the latter throughout this work.  


    We are now ready to introduce our model for shape prediction, which we will use for hippocampi.
    For this let $M$ be any shape space that is a Riemannian manifold and $U \subseteq M$ a normal convex neighborhood consisting of the shapes of interest to us. Given a shape $p^* \in U$ observed at time $t_0$ our goal is to predict its future form $q^* \in U$ at time point $t_1$. For simplicity of exposition, we assume in the following that $t_0 = 0$ and $t_1 = 1$. This can always be achieved when the times are viewed relative to an interval that contains them.      
    
    Our fundamental assumption is that $p^*$ develops along a geodesic through $U$, i.e., there is $\gamma_{p^*} \in G(U)$ such that the longitudinal development of $p^*$ at time $t$ is given by  $\gamma_{p^*}(t)$. Several works have shown that this is often an adequate choice when modeling shape developments in the medical context~\cite{Navayazdani_ea2020, PennecSommerFletcher2020}. 
    Then, since geodesics are determined by their starting point and initial velocity, we need to find $\gamma_{p^*}'(0)$; because then
    \begin{equation} \label{eq:prediction}
        q^* = \exp_{p^*}(\gamma_{p^*}'(0)).
    \end{equation}
    In other words, to predict the development of \textit{any} shape $p^* \in U$ that is of interest to us we need to approximate a vector field on $U$ that encodes the direction and speed of change that $p^*$ undergoes until time 1.  

    In the following, we propose an approach that infers this vector field from data. 
    Assume $N$ shapes similar to $p^*$ (i.e., close to $p^*$ in $U$), which are expected to show analogous progression and
    are observed at a possibly varying number of time points, hence, yielding (training) data 
    $(t^{(j)}_i, q_i^{(j)}) \in [0,1] \times U$, for $i=1,\dots,n_j$ and $j=1,\dots,N.$ 
    Now, using geodesic regression~\cite{Fletcher2013} we can approximate the individual trajectories $\gamma^{(1)}, \dots, \gamma^{(N)} \in G(U)$ of the training shapes. Utilizing the manifold structure of $G(U)$ from~\cite{hanik2022nonlinear}, the (Fr\'{e}chet) mean geodesic $\overline{\gamma}$ of $\gamma^{(1)}, \dots, \gamma^{(N)}$ can be computed. Note that $\overline{\gamma}(0)$ and $\overline{v}:= \overline{\gamma}'(0) \in T_{\overline{\gamma}(0)}M$ can be interpreted as the mean starting point and the average initial velocity of the trajectories of the training data, respectively. The geodesic $\overline{\gamma}$ is further the central fixed effect that describes the data in a geodesic hierarchical model~\cite{MF2012,nava2022hierarchical}.
    
    The fact that the training shapes are close to $p^*$ suggests that the parallel transport of $\overline{v}$ to $p^*$ is a good approximation of our target $\gamma_{p^*}'(0)$. We thus propose to use the approximation $\gamma_{p^*}'(0) \approx \Pi^{p^*}_{\overline{\gamma}(0)}(\overline{v})$ in~(\ref{eq:prediction}). Being a comparatively simple approach, our experiments show that it can be a very good choice. The processing pipeline is shown in Fig.~\ref{fig:Model}.

\section{Experiments} \label{sec:experiments}

\subsection{Data and Methodology}

\textbf{Datasets:}
    To evaluate our model we applied it to two data sets. The first was shape data of right hippocampi derived from 3D label fields provided by the Alzheimer's Disease Neuroimaging Initiative\footnote{\url{adni.loni.usc.edu}} (ADNI).
    The ADNI database contains, amongst others, 1632 brain MRI scans with segmented hippocampi. From them, we assembled three distinct groups: subjects with Alzheimer's (AD), Mild Cognitive Impairment (MCI), and cognitive normal (CN) controls. The groups contained data from 86, 201, and 116 subjects, respectively; for each subject, there were three MR images taken (approximately) six months apart. 
    Correspondence of the surfaces (2280 vertices, 4556 triangles) was established in a fully automatic manner by registering extracted isosurfaces using the functional map--based approach of~\cite{EzuzBenchen2017}. As the final preprocessing step, all meshes were aligned using generalized Procrustes analysis. 

    The second data set was taken from Dynamic FAUST~\cite{dfaust:CVPR:2017}, which is publicly available and contains the motion data of 10 subjects. The data is given as triangle meshes in correspondence (subject-wise and between subjects). For each subject, we used three meshes (6890 vertices, 13776 triangles) that constitute a raising of the left leg from the initial part of the ``one leg loose'' pattern. 

\noindent \textbf{Shape Space:}
    We used the differential coordinate model (DCM) shape space from~\cite{vonTycowicz_ea2018}. The DCM space works with triangular mesh representations and allows for explicit and fast computations.

\noindent \textbf{Comparison Methods:}
    \begin{figure*}[t]
        \begin{center}
            \includegraphics[width=\textwidth]{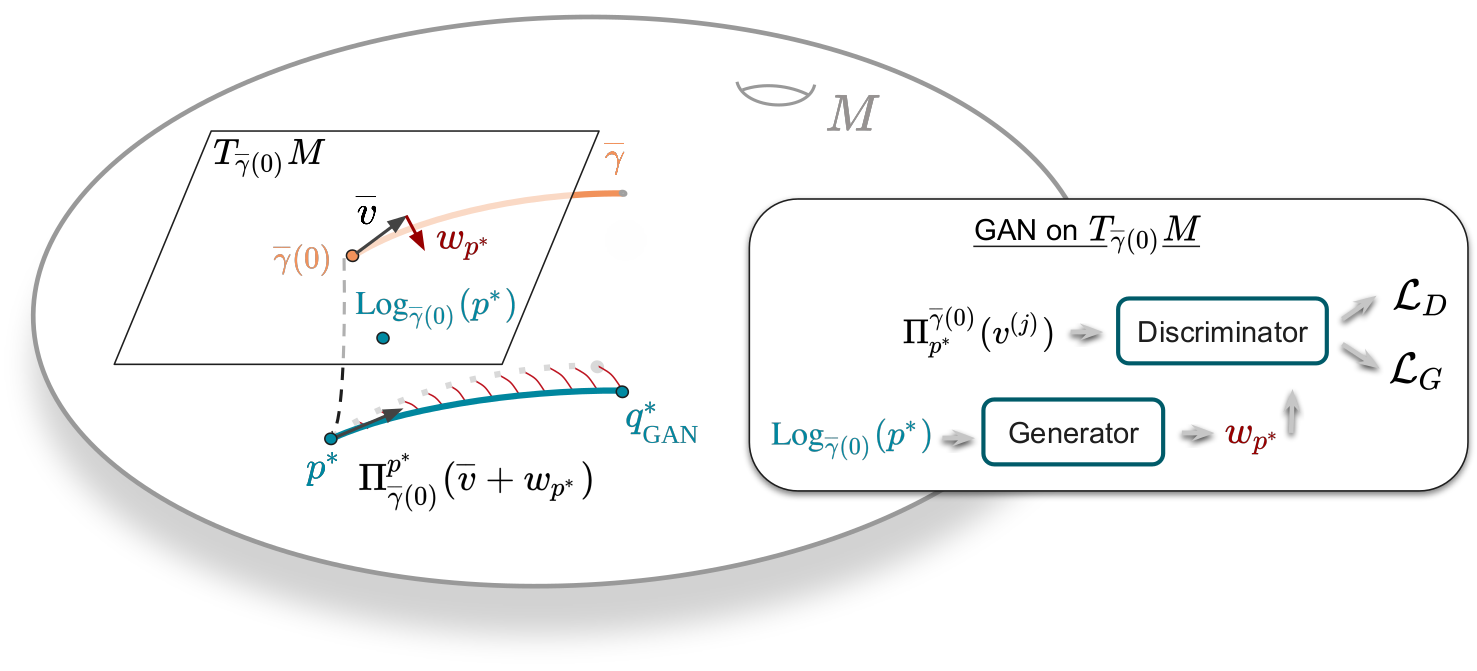}
            \caption{GAN-enhanced prediction method.}
            \label{fig:GAN_model}
        \end{center}
    \end{figure*}

    For the ADNI experiment, we used the following comparison methods.
    The first was the varifold-based (Varifold) method from~\cite{rekik2016predicting}. 
    To test whether incorporating a deep neural network improves the prediction, we also tested the following variation of our proposed method on the ADNI data:
    We tried both a generative adversarial network (GAN) and a cyclic GAN (cGAN) to learn a correction $w_{p^*} \in T_{\overline{\gamma}(0)}M$ of $\overline{v}$. The idea was that shapes from different regions in $U$ might show systematic differences in their development. Both the GAN and cGAN are designed to map the coordinate vector\footnote{We denote the \textit{coordinate representation} in $\mathbb{R}^d$ of a tangent vector $v$ w.r.t.\ a fixed but arbitrary basis by $[v]$.} of the cross-sectional difference $[\log_{\overline{\gamma}(0)}(p^*)] \in \mathbb{R}^d$ to the correction vector $[w_{p^*}] \in \mathbb{R}^d$. They were trained on the baselines $\{ q_1^{(j)}\ |\ j = 1,\dots,N\}$ using three-fold cross validation.
    We then used
    $\gamma_{p^*}'(0) \approx \Pi^{p^*}_{\overline{\gamma}(0)}(\overline{v} + w_{p^*})$ in~(\ref{eq:prediction}), i.e., the prediction then became $$q^*_{\textnormal{GAN}} := \exp_{p^*}(\Pi^{p^*}_{\overline{\gamma}(0)}(\overline{v} + w_{p^*})).$$ The method using the GAN is illustrated in Fig.~\ref{fig:GAN_model}.
    
    The generators in the GAN and cGAN networks consisted of four linear layers, with dropout layers and ReLU activation functions between them; the discriminators were composed of two linear layers without dropout. 
    Both the GAN and cGAN 
    used the sum of the binary cross entropy and the difference
    $\| \Pi_{\gamma^{(j)}(0)}^{\overline{\gamma}(0)}(v^{(j)}) - \overline{v} - w_{p^*}\|_{\overline{\gamma}(0)}$ (with the norm that is induced by the Riemannian metric) as loss; for the cGAN the standard forward cycle consistency loss was also added. When referencing results obtained with the GAN and cGAN while using the DCM space as shape space, we use the notations (DCM+GAN) and (DCM+cGAN), respectively.
    
    To evaluate how important the DCM space is to the prediction, we replaced it with the flat point distribution model (PDM) space from~\cite{Cootes_ea1995}.
    
    Finally, to further differentiate the behavior of our proposed and the varifold-based approach, we tested how well the regressed geodesics in the DCM space and the space of diffeomorphisms approximate the data. We report for all groups the averages of the mean vertex-wise error (MVE) between the mesh belonging to $\gamma^{(j)}(1)$ and the corresponding data mesh of $q^{(j)}_2$ after rigid alignment.

    In the experiment with the dynamic FAUST data, we compared our method against the varifold-based method. We did not use these deep-learning refinements for the FAUST data since only ten subjects were available. (Note that an advantage of our approach is also that it can handle such small datasets.)
    
\noindent \textbf{Software:}
    The computations in the DCM and PDM space were performed in Morphomatics v1.1~\cite{Morphomatics}. The GAN and cGAN were implemented in PyTorch.
    During training, we used ADAM optimizer for both the generators and the discriminators. 
    All computations involving varifolds were performed in Deformetrica 4.3.0rc0~\cite{bone2018deformetrica}, where we used the ``landmarks'' option as attachment type for a fair comparison since all meshes were in correspondence. In the ADNI experiment, we used the L-BFGS algorithm in Deformetrica, and gradient ascent or the Faust data as L-BFGS had stability problems there.
    
\noindent \textbf{Parameter Settings:}
    In the ADNI experiment (where measurements are in millimeters), we used Varifold with a smoothing kernel width of $2.5$, a deformation kernel width of $5$, $t_0 = 0.5$, a concentration of time points of 1, and the number of points used from the cloud was set to $25$.
    The GAN and cGAN used a dropout probability of 0.2; the learning rates of both the generators and discriminators were set to 0.0001. For each fold, we trained the network with 400 epochs.
    In the FAUST experiment (where measurements are in meters), we used Varifold with a smoothing kernel width of $0.1$, a deformation kernel width of $0.1$, $t_0 = 0.5$, a concentration of time points of 1; the number of points used from the cloud was also set to $25$.
    
    All parameters were determined through exploration. All results were obtained through three-fold cross-validation.
    
\noindent \textbf{Evaluation Measures:}
    To assess the effectiveness of our model, we employed the MVE by calculating how much (after rigid alignment) each vertex deviates from its corresponding ground truth vertex.
    Moreover, to compare the DCM and PDM spaces, we contrasted the MVE (which can be viewed as the intrinsic distance of PDM space) with the geodesic distance $d$ (GD) of DCM space.

\begin{table*}[t]
    \centering
    \caption{Comparison of (variants of) our and the varifold-based prediction method w.r.t.\ average mean vertex-wise error (MVE) and average geodesic distance (GD). 
    }
    \begin{tabular}{ c | c | c c c c c} 
         Metric & Group & DCM & PDM & DCM+GAN & DCM+cGAN & Varifold \\
        \hline
          \multirow{3}{*}{\rotatebox[origin=c]{90}{MVE}}
          & AD & $\mathbf{0.70 \pm 0.07}$ & $0.71 \pm 0.02$ & $0.73 \pm 0.02$ & $0.75 \pm 0.02$ & $1.23 \pm 0.14$ \\
          & MCI & $\mathbf{0.64} \pm 0.05$ & $\mathbf{0.64 \pm 0.03}$ & $0.67 \pm 0.03$ & $ 0.75 \pm 0.04$ & $1.27 \pm 0.02$ \\
          & CN &  $\mathbf{0.71 \pm 0.01}$ & $0.72 \pm 0.08$ & $0.74 \pm 0.07$& $0.80 \pm 0.05$ & $1.33 \pm 0.06$ \\
        \hline
          \multirow{3}{*}{\rotatebox[origin=c]{90}{GD}}
          & AD & $\mathbf{15.51 \pm 0.88}$ & $37.50 \pm 0.79$ & $ 16.20 \pm 0.60$ & $16.42 \pm 0.47$ & $17.63 \pm 1.97$ \\
          & MCI & $\mathbf{15.23 \pm 0.87}$ & $34.55 \pm 1.81$ & $15.91 \pm 0.89$ & $16.22 \pm 0.50$ & $18.36 \pm 0.25$ \\
          & CN & $\mathbf{16.44 \pm 0.22}$ & $38.39 \pm 4.44$ & $17.35 \pm 1.66$ & $18.35 \pm 1.27$ & $19.07 \pm 1.13$ \\
        \hline 
    \end{tabular}
    \label{tab:table_1}
\end{table*}

\subsection{Results} \label{sec:Results}

    \begin{table*}[b] 
    \begin{small}
        \caption{Regression fidelity in terms of MVE.}
        \begin{center}
            \begin{tabular}{ c | c c c} 
            \hline
             Group & AD & MCI & CN \\
            \hline
              DCM  &  $\mathbf{0.16 \pm 0.06}$ & $\mathbf{0.17 \pm 0.16}$ & $\mathbf{0.18 \pm 0.18}$ \\
              Varifold  &  $0.68 \pm 0.58$ & $0.62 \pm 0.47$ & $0.73 \pm 0.74$ \\
            \hline 
            \end{tabular}
            \label{tab:regression_fidelity}
        \end{center}
    \end{small}
    \end{table*} 

    \begin{figure}[tb]
        \centering
        \includegraphics[width=\textwidth]{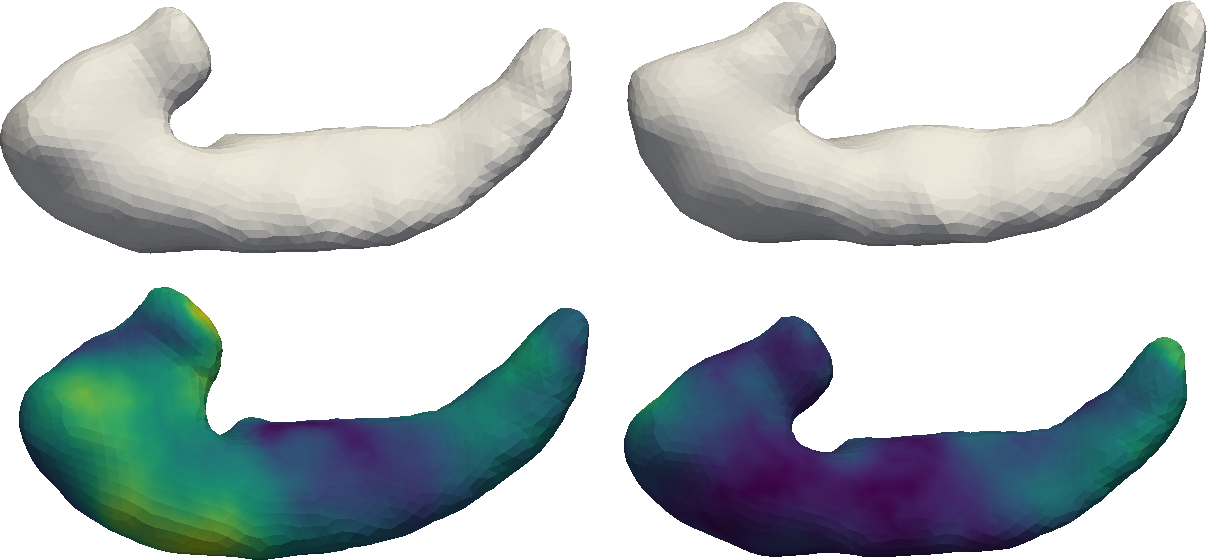}
        \vspace{-1em}
        \caption{Prediction comparison for a hippocampus from the AD group. Upper left: base shape at $t=0$; upper right: the same hippocampus at $t=1$, i.e., after one year (the ground truth for our prediction); bottom left: varifold-based prediction; bottom right: our prediction. The colors encode the vertex-wise differences to the ground truth according to the color map (0 mm~\ShowColormap~2.25 mm).}
        \label{fig:hippocampi}
        \vspace{-1em}
    \end{figure}
    
    The results of our prediction comparison for the hippocampi are shown in Table~\ref{tab:table_1}. Our proposed method outperformed the other methods in all categories.
    Even though the proposed method is a relatively simple approach, it not only performed better than GAN methods but was also faster and did not involve hyperparameter tuning. Also, differently structured GANs did not improve the results by a noticeable amount. Furthermore, we can see that using the DCM space is superior to using the PDM space as its results are close w.r.t.\ MVE (the intrinsic PDM distance) but significantly worse w.r.t.\ the GD of the DCM space. Note that the magnitude of the prediction errors of our method is probably close to the resolution of the scanner. However, the longitudinal changes in shape over the course of one year are also only small.
    
    In comparison to the varifold-based approach, the proposed method achieves superior results. 
    The difference in prediction is visualized in Fig.~\ref{fig:hippocampi} using a hippocampus from the AD group for which the MVE of our method was very close ($\approx 0.69$) to its mean MVE. The MVE of the varifold-based method was slightly smaller ($\approx 1.03$) than its MVE. 
    
    A reason why our approach works better than the varifold-based one could be the difference in approximation power of geodesic regression: Results for the fitting quality of regression in DCM and the space of diffeomorphisms are shown in Table~\ref{tab:regression_fidelity}. Clearly, the approximation is better when the DCM space is used, thus, demonstrating an improved fidelity of DCM geodesics over diffeomorphic representations.

\begin{figure}[tb]
    \centering
    \includegraphics[width=\textwidth]{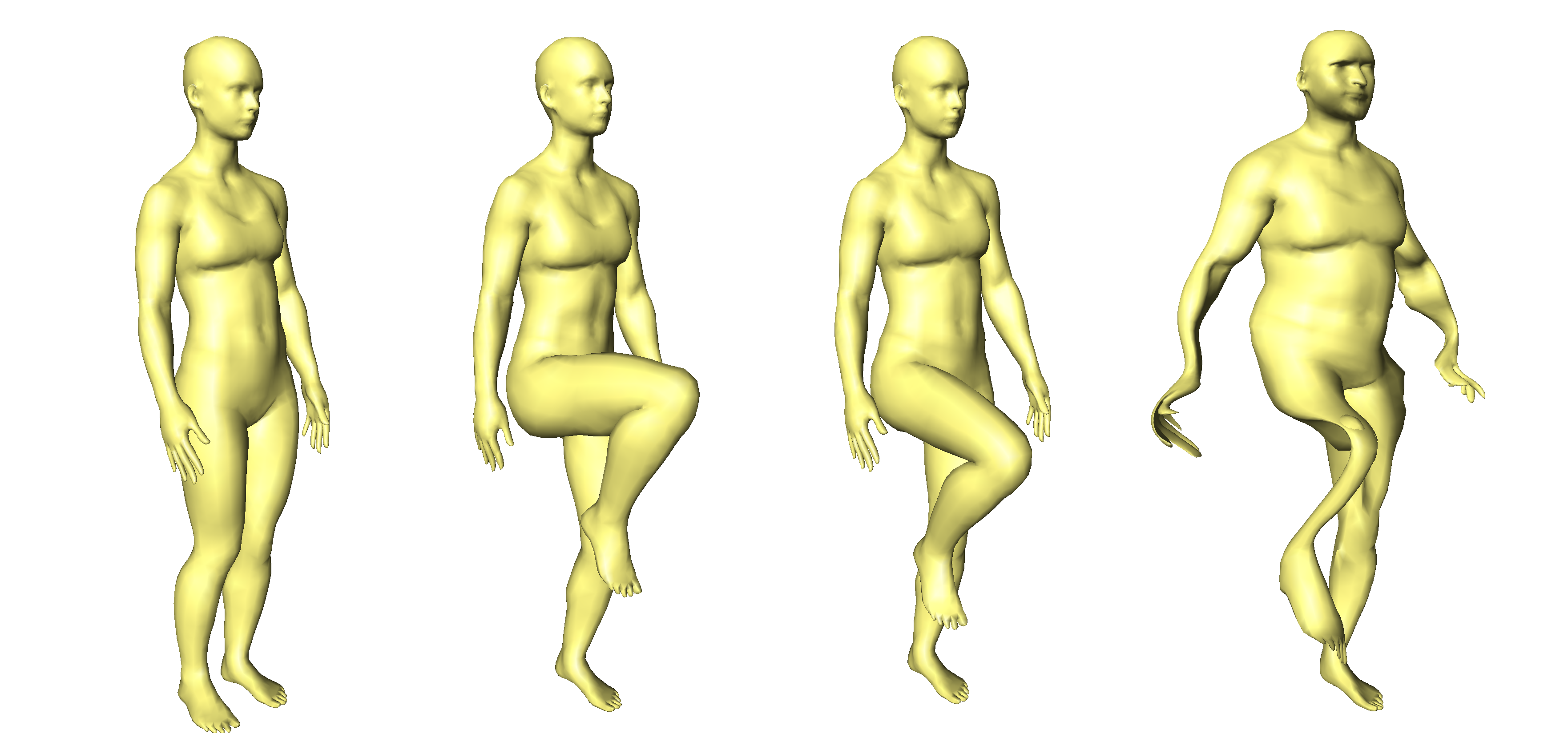}
    \vspace{-1em}
    \caption{Prediction of human body motion for FAUST dataset. From left to right: Baseline, ground truth, and predicted shapes using our and the Varifold method~\cite{rekik2016predicting}.}
    \label{fig:FAUST}
    \vspace{-1em}
\end{figure}

    The results for the FAUST dataset reinforce the above findings. Our proposed method (DCM) achieves an MVE of $0.03 \pm 0.003$ and a GD of $0.54 \pm 0.03$; Varifold has an MVE of $0.07 \pm 0.008$ and a GD of $1.16 \pm 0.04$. Fig.~\ref{fig:FAUST} shows the predictions for subject 500021. While the result of our method is close to the ground truth, the human is barely recognizable in the varifold-based prediction. A reason is the high inter- and intra-subject variability in the motion. Since point trajectories from different subjects are mixed to produce the varifold prediction, relatively unlikely leg configurations are obtained.
    Moreover, the formation of a male face in the varifold case highlights the advantage of a differential encoding of shape changes (i.e.\ tangent vectors) together with a consistent transport over combinations of absolute configurations.  

\section{Conclusion} \label{sec:conclusion}
    In this paper, we proposed a novel method for predicting shape development based on hierarchical statistical modeling in Riemannian shape spaces. It outperformed state-of-the-art in two experiments by a clear margin. Furthermore, it performed better than deep learning--supported variants when predicting the future development of the hippocampus shape. Our approach is thus a good fit for shapes whose progression follows geodesics (such as hippocampi) and who are well captured by population-average trends. 
    As the latter assumption will not always be valid,
    it is still a promising approach to incorporate deep learning: Whenever the development depends strongly on the individual characteristics of the baseline shape, we expect that deep learning methods can be used to find adjustments to our prediction (direction) that take the dependence into account. A path for future work is thus to test this hypothesis on further anatomical structures.  

\subsubsection{Acknowledgements} \label{sec:acknowledgments}
This work was partially funded by grants from the European H2020 Marie Sklodowska-Curie action (grant no.\ 101003403) and the Scientific and Technological Research Council of Turkey under the TUBITAK 2232 Fellowship for Outstanding Researchers (no.\ 118C288)\footnote{\href{https://basira-lab.com/normnets/}{https://basira-lab.com/normnets/} \& \href{https://basira-lab.com/reprime/}{https://basira-lab.com/reprime/}}.
We are grateful for the funding by DFG\footnote{Deutsche Forschungsgemeinschaft (DFG) through Germany’s Excellence Strategy – The Berlin Mathematics Research Center MATH+ (EXC-2046/1, project ID: 390685689)} and BMBF\footnote{Bundesministerium f\"ur Bildung und Forschung (BMBF) through BIFOLD - The Berlin Institute for the Foundations of Learning and Data (ref. 01IS18025A and ref 01IS18037A)}. 

Data collection and sharing for this project was funded by the ADNI\footnote{\href{https://adni.loni.usc.edu}{adni.loni.usc.edu}} (National Institutes of Health Grant U01 AG024904) and DOD ADNI (Department of Defense award number W81XWH-12-2-0012). ADNI is funded by the National Institute on Aging, the National Institute of Biomedical Imaging and Bioengineering, and through generous contributions from the following: AbbVie, Alzheimer's Association; Alzheimer's Drug Discovery Foundation; Araclon Biotech; BioClinica, Inc.; Biogen; Bristol-Myers Squibb Company; CereSpir, Inc.; Cogstate; Eisai Inc.; Elan Pharmaceuticals, Inc.; Eli Lilly and Company; EuroImmun; F. Hoffmann-La Roche Ltd and its affiliated company Genentech, Inc.; Fujirebio; GE Healthcare; IXICO Ltd.; Janssen Alzheimer Immunotherapy Research \& Development, LLC.; Johnson \& Johnson Pharmaceutical Research \& Development LLC.; Lumosity; Lundbeck; Merck \& Co., Inc.; Meso Scale Diagnostics, LLC.; NeuroRx Research; Neurotrack Technologies; Novartis Pharmaceuticals Corporation; Pfizer Inc.; Piramal Imaging; Servier; Takeda Pharmaceutical Company; and Transition Therapeutics. The Canadian Institutes of Health Research is providing funds to support ADNI clinical sites in Canada. Private sector contributions are facilitated by the Foundation for the National Institutes of Health (www.fnih.org). The grantee organization is the Northern California Institute for Research and Education, and the study is coordinated by the Alzheimer's Therapeutic Research Institute at the University of Southern California. ADNI data are disseminated by the Laboratory for Neuro Imaging at the University of Southern California.



\bibliographystyle{splncs04}
\bibliography{refs}

\end{document}